\def\year{2022}\relax
\documentclass[letterpaper]{article} 
\usepackage{aaai22}  
\usepackage{times}  
\usepackage{helvet}  
\usepackage{courier}  
\usepackage[hyphens]{url}  
\usepackage{graphicx} 
\usepackage{natbib}  
\usepackage{caption} 
\DeclareCaptionStyle{ruled}{labelfont=normalfont,labelsep=colon,strut=off} 
\usepackage{algorithm}
\usepackage{algorithmic}
\usepackage{xcolor}
\usepackage{amsmath}
\usepackage{booktabs}
\usepackage{multirow}

\usepackage{newfloat}
\usepackage{listings}
\floatstyle{ruled}
\newfloat{listing}{tb}{lst}{}
\floatname{listing}{Listing}
\nocopyright
\title{Impact of Subword Pooling Strategy on Cross-lingual Event Detection}
\author{
    Shantanu Agarwal\equalcontrib,
    Steven Fincke\equalcontrib,
    Chris Jenkins,
    Scott Miller,
    Elizabeth Boschee
}
\affiliations{
    University of Southern California Information Sciences Institute\\
    \{shantanu, sfincke, cjenkins, smiller, boschee\}@isi.edu
}

\usepackage{bibentry}

\begin{document}

\maketitle

\begin{abstract}
Pre-trained multilingual language models (e.g., mBERT, XLM-RoBERTa) have significantly advanced the state-of-the-art for zero-shot cross-lingual information extraction.
These language models ubiquitously rely on word segmentation techniques that break a word into smaller constituent subwords.
Therefore, all word labeling tasks (e.g. named entity recognition, event detection, etc.), necessitate a pooling strategy that takes the subword representations as input and outputs a representation for the entire word.
Taking the task of cross-lingual event detection as a motivating example, we show that the choice of pooling strategy can have a significant impact on the target language performance.
For example, the performance varies by up to 16 absolute $f_{1}$ points depending on the pooling strategy when training in English and testing in Arabic on the ACE task. 
We carry out our analysis with five different pooling strategies across nine languages in diverse multi-lingual datasets.
Across configurations, we find that the canonical strategy of taking just the first subword to represent the entire word is usually sub-optimal. 
On the other hand, we show that attention pooling is robust to language and dataset variations by being either the best or close to the optimal strategy.
For reproducibility, we make our code available at https://github.com/isi-boston/ed-pooling.
\end{abstract}

\section{Introduction}

The goal of event detection (ED) is to identify instances of real-world events in text. Each event consists of a phrase which evokes the event, along with a classification of the event into a pre-specified ontology of event types. 
For example, given a sentence \textit{Mathias Cormann took office last week}, an event detection system for the ACE \cite{doddington-etal-2004-automatic} ontology needs to infer that the phrase \textit{took office} is an event of type \textit{Start-Position}.
ED is a knowledge extraction step that has been shown to be critical for many NLP applications such as question answering \cite{10.1145/860435.860444}, information retrieval \cite{Basile2014ExtendingAI}, argument extraction \cite{cheng-erk-2018-implicit}, etc.

Current ED systems rely on neural network architectures that require large amounts of data for training.
Getting high-quality training data for a structured task like ED is challenging and expensive.
Because of this, most of the training datasets are available only in English.
Bridging the ED performance gap between English and other non-English (likely low-resource) languages is an active area of research (e.g.\ \citet{haoran-etal-2022-gradual-finetuning}, \citet{fincke2022}, \citet{huang-etal-2022-multilingual-generative}).

Most of the current cross-lingual ED systems leverage pre-trained multilingual language models such as
mBERT \cite{devlin-etal-2019-bert} or XLM-RoBERTa \cite{conneau-etal-2020-unsupervised}.
These models rely on data-driven word segmentation techniques such as BPE \cite{sennrich-etal-2016-neural}, WordPiece \cite{DBLP:journals/corr/WuSCLNMKCGMKSJL16}, SentencePiece \cite{kudo-richardson-2018-sentencepiece}, etc.
Working with segmented words (subwords) enables models to effectively encode rare words, thus mitigating out-of-vocabulary issues.

However, using subwords can pose challenges, particularly for word labeling tasks such as event detection. 
Here, to form a representation for each original word, we must find a way to combine the representations produced by the language model for each subword. 
However, coming up with a pooling strategy is challenging.
This is particularly true in a zero-shot cross-lingual context, where simple heuristics (e.g., just use the representation for the first subword) are less likely to be applicable for both the source and target languages. 

In English, for instance, that ``first subword'' heuristic may be quite reasonable for event detection, since the first subword typically carries the most information about the meaning of the entire word. 
As an example, XLM-RoBERTa tokenizes the English word 
\textit{attacked} to \textit{attack ed}\footnote{The `\_' marker is removed for simplicity.}, with the first subword \textit{attack} likely being far more useful for classification than \textit{ed} would be.
However, in Arabic, the reverse is often true. The word
\textit{whAjm}\footnote{Buckwalter transliteration is provided for Arabic.} (translated as ``and (he) attacked') gets tokenized as \textit{w hAjm}, with the subwords meaning ``and'' and ``(he) attacked''.
Here, it is the last subword that carries the most useful meaning for classification and if an ED model were to consider only the representation for the first subword, it might very likely fail to correctly classify this word as an event.

In this paper, we explore the impact of different subword-to-word pooling strategies on the zero-shot cross-lingual event detection task.
Our main contributions are:
\begin{itemize}
    \item{When using massively multilingual models such as XLM-RoBERTa, we show that the choice of the pooling strategy can have a significant impact on the performance of cross-lingual event detection.}
    \item{Across a diverse set of languages, we show that attention-pooling works best and that the canonical strategy of first-subword pooling can be sub-optimal.
    }
    \item{When using bilingual models, i.e. models that only need to share vocabulary across the source (English) and one target language, the cross-lingual performance is less sensitive to the pooling strategy.}
\end{itemize}

\section{Related work}
There is a growing body of work showing how statistical word segmentation methods adversely affect the performance of pretrained language models when dealing with morphologically rich languages like Arabic, Hebrew, Turkish, etc. (\citet{amrhein-sennrich-2021-suitable-subword},  \citet{https://doi.org/10.48550/arxiv.2204.04748}).
Instead of being completely data-driven, these studies advocate for subword tokenization techniques to be linguistically motivated such that the subwords adhere to morpheme boundaries.
Although appealing, coming up with bespoke language-dependent tokenization which scales to a multitude of languages (of the order of 100 as in mBERT/XLM-RoBERTa) is impractical and prohibitively laborious.

Very few studies have looked into the impact of pooling strategies on word labeling tasks.
\citet{toshniwal-etal-2020-cross} studied the impact of span representation of text for `edge-probing' tasks such as constituent labeling, named entity labeling, semantic role labeling, and coreference arc prediction.
However, their analysis solely focused on English.
Closest to our work is the analysis by  \citet{acs-etal-2021-subword}.
They studied the effect of subword pooling on three tasks: morphological probing, POS tagging and NER.
Although their analysis covered languages other than English, they solely focused on the monolingual setting.
Our work differs in that we probe the effect of subword pooling not on monolingual scenarios but for zero-shot cross-lingual extraction.
 
\section{Datasets}
We probe the impact of subword pooling on the following diverse set of cross-lingual event detection datasets:

\textit{\textbf{BETTER Abstract}}:
Better Extraction from Text Towards Enhanced Retrieval (BETTER) is an information extraction and information retrieval program\footnote{https://ir.nist.gov/better}.
One of the tasks in this program is Abstract event extraction.
For this task, training (and validation) data is available only in English and test data is available in English, Arabic, Farsi and Korean.
There are no event type distinctions in the Abstract event task\footnote{Following the practice of the BETTER program, we drop the quad-class information from the Abstract dataset.}.

\textit{\textbf{BETTER Basic Phase-1}}:
In contrast to the Abstract task, the events in the BETTER Basic task have event types associated with them.
The Phase-1 ontology has 69 event types. The training (and validation) annotations are provided for English and test data is available for English and Arabic.

\textit{\textbf{BETTER Basic Phase-2}}:
The Phase-2 BETTER Basic task expands the Phase-1 ontology to cover 92 event types.
Training (and validation) annotations are provided for English and test data is available for English and Farsi.

\textit{\textbf{ACE}}:
The Automatic Content Extraction (ACE) dataset \cite{doddington-etal-2004-automatic} is extensively used in literature to benchmark monolingual and cross-lingual capabilities of information extraction systems.
For this task, training, validation, and test data is available in English, Arabic and Chinese.
Since Chinese is a non-white-space delimited language, we have excluded it from the present effort.
To be consistent with previous works (\citet{haoran-etal-2022-gradual-finetuning}, \citet{fincke2022}, \citet{huang-etal-2022-multilingual-generative}), we use the same train/dev/test splits as used in \cite{huang-etal-2022-multilingual-generative}.

\textit{\textbf{MINION}}:
An event detection dataset spanning eight different languages was introduced by \citet{pouran-ben-veyseh-etal-2022-minion}.
The MINION dataset borrows its ontology from ACE and has 16 event types.
We use the official train/dev/test splits to report our scores.
We exclude Japanese from our analysis as it is a non-white-space delimited language.

\begin{table*}[hbt!]
\centering
\begin{tabular}{l l c c c c c c}
\toprule
Task & Lang. & First subword & Last subword & Average & IDF & Attention \\ \midrule\midrule
\multirow{4}{*}{ BETTER Abstract }
  & en   & 87.8             & \underline{87.6} & \textbf{88.0}    & 87.7          & 87.7            \\
  & ar   & \underline{59.6} & 74.1             & 67.9             & 76.4          & \textbf{78.1}   \\
  & fa   & \textbf{73.9}    & \underline{66.1} & 69.0             & 70.6          & 70.4             \\
  & ko   & 75.9             & \underline{55.4} & 67.1             & 72.8          & \textbf{79.1}    \\
\midrule
\multirow{2}{*}{ BETTER Phase-1 }
  & en   & 66.2             & \textbf{67.0}    & 66.1             & 66.1          & \underline{64.7} \\
  & ar   & \underline{48.2} & 55.3             & 51.0             & \textbf{56.7} & 54.6             \\
\midrule
\multirow{2}{*}{ BETTER Phase-2 }
  & en   & 65.8             & \textbf{66.2}    & 65.7             & \textbf{66.2} & \underline{65.6} \\
  & fa   & \textbf{53.4}    & \underline{50.6} & 51.3             & 53.3          & \textbf{53.4}   \\
\midrule
\multirow{2}{*}{ ACE }
  & en   & 70.2             & 70.2             & \underline{69.9} & \textbf{71.5}  & 71.3            \\
  & ar   & \underline{42.0} & 53.3             & 44.3             & 57.9           & \textbf{58.0}  \\
\midrule
\multirow{7}{*}{ MINION }
  & en   & \underline{79.3} & 79.5            & \underline{79.3}  & 79.4           & \textbf{79.7}   \\
  & es   & 64.5             & \underline{63.6}& 64.2              & \textbf{65.0}  & 63.7            \\
  & pt   & 75.3             & 74.7            & 74.4              & \textbf{75.8}  & \underline{74.2} \\
  & pl   & 64.2             & \underline{62.2}& \textbf{64.3}     & \textbf{64.3}  & 64.0             \\
  & tr   & \textbf{59.5}    & \underline{47.9}& 52.2              & 55.5           & 57.8             \\
  & hi   & 72.3             & \underline{69.9}& 70.4              & 72.3           & \textbf{73.1}    \\
  & ko   & \textbf{65.4}    & \underline{40.4}& 52.9              & 54.2           & 63.1            \\
\midrule
\midrule
\multirow{1}{*}{ Avg. }
& ---    & 66.0             & \underline{63.7}& 64.5              & 67.3           & \textbf{68.1}    \\
\bottomrule

\end{tabular}
\caption{Variation of zero-shot event detection performance across different subword pooling strategies. For a given task and language (i.e. for a particular row), we boldface the highest score and underline the lowest one. All experiments use XLM-RoBERTa large and are trained only on the English training split corresponding to the task. We report the mean $f_{1}$ score calculated over 4 runs initialized with different random seeds.}
\label{tab:main}
\end{table*}

\section{Architecture and pooling strategies}
Suppose we are given a sentence $S$ which, using some word-based tokenizer (spaCy, nltk, etc.), is tokenized to a sequence of words $[w_0, w_1, \dots, w_n]$.
Then, each word is further passed through the language model specific tokenizer and gets broken into subwords, i.e. $w_i \rightarrow [t^{i}_0, t^{i}_1, \dots, t^{i}_m]$.
The embedding for each of the subwords is passed to the transformer and we use the hidden state of the last transformer layer, $h^{i}_{j}$, as the representation for $t^{i}_{j}$.
A pooling function $f$ takes the hidden states of the subwords and outputs a fixed-dimensional representation, $f(h^{i}_0, \dots, h^{i}_m)$, for the word $w_i$.
The pooled representation, $f(h^{i}_0, \dots, h^{i}_m)$ is then passed through a single layer feed-forward network to classify the word to be either the beginning (B), inside (I) or not (O) a part of an event span.

The focus of this paper is on the choice of pooling function $f$. Below we describe five different pooling functions which we explore as part of our event detection system.

\textbf{First subword pooling:}
We simply take the representation of the first subword as a proxy to represent the whole word:
\begin{equation}
    f(h^{i}_0, \dots, h^{i}_m) = h^{i}_0.
\end{equation}
This is the default trigger classification configuration which prior works on event detection typically use, e.g. \citet{fincke2022}. 
This is also the canonical approach used by Huggingface\footnote{https://github.com/huggingface}, a widely used framework, for benchmarking sequence labeling tasks.

\textbf{Last subword pooling:}
Take the representation of the last subword as a proxy to represent the whole word:
\begin{equation}
    f(h^{i}_0, \dots, h^{i}_m) = h^{i}_m.
\end{equation}

\textbf{Average pooling:}
The word representation is the mean of all the hidden representations:
\begin{equation}
    f(h^{i}_0, \dots, h^{i}_m) = \frac{1}{m + 1} \sum_{j=0}^{m}h^{i}_{j}.
\end{equation}

\textbf{IDF pooling:}
A pooling strategy should ideally assign the most weight to the subwords which will best inform the classifier in its decision making process. Because average pooling fails to do so and puts equal importance on each subword, we hypothesize that average pooling is a sub-optimal strategy.

One way to estimate the meaningful information content of a subword is by the frequency at which one observes the subword in a corpus; less frequent subwords have more information than subwords which occur frequently.
We quantify this notion by assigning each subword, $t$, its inverse document frequency (IDF) score:
\begin{equation}
    idf(t) = \log{\frac{|W|}{|w:t\in w|}},
\end{equation}
where $|W|$ is the total number of words in a corpus and $|w:t\in w|$ is the number of words which contain $t$ as a subword.
For pooling, the weights assigned to the subwords are determined by normalizing these $idf$ scores through a softmax:
\begin{equation}
\begin{aligned}
    &a_j = \text{softmax}(\bf{idf})_j \\
    &f(h^{i}_0, \dots, h^{i}_m) = \sum_{j=0}^{m} a_j h^{i}_j.
\end{aligned}
\end{equation}

\textbf{Attention pooling:}
There are a few drawbacks to the IDF pooling strategy: 1) information we want to extract from a word is task-dependent and IDF may or may not be a good quantitative measure to capture that, 2) the technique relies on calculating the IDF scores on a (large) corpus, thus coupling the system performance to the corpus statistics and 3) at test time, one can encounter subwords which are not seen when preparing IDF scores, thus requiring an additional hyper-parameter, i.e. the default IDF score for such unseen subwords.

All these limitations can be mitigated through attention pooling where a \textbf{learnt} query vector, $v$, is used to determine the weight that needs to be assigned to each hidden representation:
\begin{equation}
\begin{aligned}
    \alpha_j &= v \cdot h^{i}_j; \; a_j = \text{softmax}(\mathbf{\alpha})_j \\
    f(h^{i}_0, \dots, h^{i}_m) &= \sum_{j=0}^{m} a_j h^{i}_j.
\end{aligned}
\end{equation}

\section{Results and Analysis}
In Table \ref{tab:main}, we show the zero-shot event detection performance for all datasets and languages using different subword pooling strategies. 
For all configurations, we train our models on the English data and use XLM-RoBERTa large as our pre-trained language model. 
We ensure that the same pooling strategy is used for training and testing. 
We make the following observations:

\textbf{Attention pooling is usually the best or close to the optimal strategy:}
By picking a pooling strategy, we impose our prior beliefs about which subwords we think have the most information content. 
For example, when using \textit{First subword} pooling, we are assuming that the first subword has the most information necessary for the model to classify whether the word is in an event span or not. 
Depending on the language of interest, this constraint may or may not be correct and can result in sub-optimal system performance, especially in the zero-shot cross-lingual scenario where the language structure of the source and target languages can be markedly different.  

In Table \ref{tab:main}, we see that attention pooling is either the best or close to being optimal.  
This can be attributed to the fact that across all the pooling strategies, attention pooling has the least inductive bias; the process of finding which subword is important is learnt in an end-to-end manner and no a-priori constraints (either linguistic as in first/last subword or corpus driven as in IDF) are imposed. 

\textbf{Variation across pooling strategies is higher for languages with high shattering rate:}
When pre-training a massively multilingual model such as XLM-RoBERTa, not all languages are represented equally. 
This non-uniform language representation often leads to words from low-resource languages (such as Arabic) to split more often than words from high-resource languages (such as English).
Shattering rate quantifies this by calculating the average number of subwords a word is likely to split into.
In Table \ref{tab:shattering}, we show the shattering rate of the anchor words for each of the target languages in our experiments along with how variable the target language performance is across various pooling strategies. 
We quantify variation ($\Delta$) as the difference between the scores for the best and the worst pooling strategy. 
From Table \ref{tab:shattering}, we see that high variability is observed with high shattering rates.

\begin{table}[hbt!]
\small
\centering
\begin{tabular}{l l c c }
\toprule
Task & Lang. & Shattering rate & $\Delta$ \\ \midrule\midrule
\multirow{4}{*}{ BETTER Abstract }
  & en   & 1.5   & 0.4  \\
  & ar   & 2.0   & 18.5 \\
  & fa   & 1.2   & 7.8  \\
  & ko   & 2.4   & 23.7 \\
\midrule
\multirow{2}{*}{ BETTER Phase-1 }
  & en   & 1.6   & 2.3  \\
  & ar   & 2.0   & 8.5  \\
\midrule 
\multirow{2}{*}{ BETTER Phase-2 }
  & en   & 1.5   & 0.6  \\
  & fa   & 1.3   & 2.8  \\
\midrule 
\multirow{2}{*}{ ACE }
  & en   & 1.5   & 1.6  \\
  & ar   & 2.1   & 16.0 \\
\midrule 
\multirow{7}{*}{ MINION }
  & en   & 1.4   & 0.4  \\
  & es   & 1.6   & 1.4  \\
  & pt   & 1.7   & 1.6  \\
  & pl   & 2.1   & 2.1  \\
  & tr   & 1.7   & 11.6 \\
  & hi   & 1.3   & 3.2  \\
  & ko   & 2.6   & 25.0 \\
\bottomrule

\end{tabular}
\caption{Shattering rate of anchor words in the target language dataset and sensitivity ($\Delta$) to pooling strategies.}
\label{tab:shattering}
\end{table}

\begin{table*}[hbt]
\centering
\begin{tabular}{l c c c c c c c}
\toprule
Model            & Shatt. (En/Ar) & First subword      & Last subword & Average          & IDF  & Attention     & $\Delta$ \\ \midrule\midrule

XLM-RoBERTa      & 1.5/2.0             & \underline{59.6} & 74.1       & 67.9             & 76.4 & \textbf{78.1} & 18.5     \\
GigaBERT         & 1.1/1.3             & 62.6             & 63.0       & \underline{61.8} & 65.4 & \textbf{65.7} & 3.9      \\
\bottomrule

\end{tabular}
\caption{
BETTER Abstract event detection performance across pooling strategies. 
For all experiments, we train our models on English and evaluate on Arabic. 
The table provides a comparison between XLM-RoBERTa and GigaBERT.
We report mean $f_{1}$ scores averaged over four randomly initialized seeds. 
Highest scores are in bold and the lowest are underlined.
The last column indicates the variation ($\Delta$) across the pooling strategies.
The shattering rate for the source and target language (en and ar) for XLM-RoBERTa and GigaBERT is provided in the second column.
}
\label{tab:giga}
\end{table*}

We further confirm this correlation by experimenting with a language model that rarely splits words in either the source or target language, specifically the bilingual GigaBERT\footnote{GigaBERT-v4-Arabic-and-English} \cite{Lan2020GigaBERTZT}. 
GigaBERT is pre-trained only on English and Arabic and has much smaller shattering rates of 1.1/1.3 for English/Arabic as compared to 1.5/2.0 for XLM-RoBERTa. 
We see from Table \ref{tab:giga} that although GigaBERT is not as competitive as XLM-RoBERTa\footnote{Most likely because GigaBERT is trained on much less English data and has a smaller model capacity than XLM-RoBERTa.}, the variation ($\Delta$) across pooling strategies for GigaBERT  is much smaller than for XLM-RoBERTa.

Still, high shattering rates alone do not always indicate high sensitivity. We see the highest variation in languages where shattering rate is high \textit{and} individual subwords frequently represent essentially independent elements of meaning. 
For instance, Arabic has many short, highly frequent subwords (e.g., conjunctions, prepositions, and the definite marker \textit{Al-}) which are often attached to content-bearing words. 
Similarly, Turkish and Korean are languages whose affixes (commonly separated as subwords by XLM-RoBERTa) vary little according to any characteristics of the stem.
For example, XLM-RoBERTa tokenizes Turkish \textit{yedim} ``I ate'' to \textit{yedi m}. 
The affix \textit{-m} can be attached to a wide variety of words (including both nouns and verbs) but will always convey a sense of the first person singular subject or possessor. 
We hypothesize that, even though the language model outputs are contextualized, the representation for these kind of ``independent'' subwords will tend to be largely independent of the stems to which they attach.
If our pooling strategy considers only the output for a word like \textit{-m} (or weights it too heavily), our task will suffer because we miss the most relevant semantic content. 

However, this is less likely to happen with languages where the morphological relationships between subwords are more entangled. 
Spanish, Portuguese and, especially, Polish exhibit rather modest sensitivity despite having elevated shattering rates. 
Affixes in these three languages generally take forms conditioned by characteristics of the stem
(e.g. the stem's conjugation class or grammatical gender). 
We hypothesize that this discourages the language model from entirely disassociating an affix from its stem, meaning that an affix is likely to bear much of the stem's meaning. 
This reduces sensitivity to pooling strategy, since all subwords carry some of the meaning necessary to make classification decisions.

\begin{table*}[hbt]
\centering
\begin{tabular}{l c c c c c c c c}
\toprule
Task & Src./Tgt. & Shatt. (Tgt.) & First subword & Last subword & Average & IDF & Attention & $\Delta$ \\ \midrule\midrule
\multirow{2}{*}{ ACE }
  & en/ar   & 2.1 & \underline{42.0} & 53.3             & 44.3          & 57.9             & \textbf{58.0} & 16.0 \\
  & ar/ar   & 2.1 & \underline{60.5} & \textbf{63.0}    & 61.1          & 62.4             & 62.5          & 2.5  \\
\midrule
\multirow{2}{*}{ MINION }
  & en/ko   & 2.6 & \textbf{65.4}    & \underline{40.4}& 52.9              & 54.2           & 63.1         & 20.0 \\
  & ko/ko   & 2.6 & 74.7             & \underline{74.5}& 77.2              & \textbf{77.8}  & 77.1         & 3.3  \\
\bottomrule

\end{tabular}
\caption{
Event detection performance across pooling strategies. 
We compare zero-shot cross-lingual case (en/ar and en/ko) with the monolingual scenario (ar/ar and ko/ko). 
All experiments use XLM-RoBERTa large. 
We report mean $f_{1}$ scores averaged over four randomly initialized seeds. 
Highest scores are in bold and the lowest are underlined.
The last column indicates the variation ($\Delta$) across the pooling strategies.
The shattering rate for the target language (ar for ACE and ko for MINION) is provided in the third column.
}
\label{tab:mono}
\end{table*}

\begin{table*}[hbt!]
\centering
\begin{tabular}{l l c c c c c c}
\toprule
Task &  Lang. & First subword & Last subword & Average & IDF & Attention & First stem \\ \midrule\midrule
{ BETTER Abstract }
  & ar & \underline{59.6}    & 74.1            & 67.9                & 76.4           & \textbf{78.1}   & 76.9  \\
{ BETTER Phase-1 }
  & ar & \underline{48.2}    & 55.3            & 51.0                & \textbf{56.7}  & 54.6            & 56.4  \\
{ ACE } 
  & ar & \underline{42.0}    & 53.3            & 44.3                & 57.9           & \textbf{58.0}   & 54.4  \\
\midrule 
\midrule
\multirow{1}{*}{ Avg. }
   & ar & \underline{49.9}   & 61.6            & 54.4                & \textbf{63.7}  & 63.6            & 62.6  \\
\bottomrule

\end{tabular}
\caption{
Event detection performance across pooling strategies, including \textit{first stem} pooling. 
All experiments use XLM-RoBERTa large and is trained on English and is evaluated on Arabic. 
We report mean $f_{1}$ scores averaged over four randomly initialized seeds. 
Highest scores are in bold and the lowest are underlined.
}
\label{tab:first_stem}
\end{table*}

\textbf{Monolingual performance is less sensitive to pooling strategies:}
From Table \ref{tab:main}, we see that when we train and test in English, the event detection performance is not very sensitive to the pooling strategies (maximum variation is around a couple of $f_{1}$ point).
This can be attributed to two orthogonal aspects: a) English is a high resource language with a low shattering rate (see Table \ref{tab:shattering}) and b) the sensitivity to the pooling strategy is remarkably low in the monolingual setting (which we show next). 

To remove the confounding effect of low shattering rate and to explicitly elucidate that pooling strategy is less important in the monolingual scenario, we train and test our event detection system on two languages which have high shattering rates: Arabic and Korean. 
The performance variation across pooling strategies for the monolingual scenario and a comparison of what we get in the zero-shot cross-lingual case (train in English and test in either Arabic or Korean) is shown in Table \ref{tab:mono}.
From the table, we see that even though Arabic and Korean have very high shattering rates, the variability in performance is markedly low in the monolingual scenario (2.5 for Arabic/Arabic and 3.3 in Korean/Korean) compared to the variability in the zero-shot cross-lingual case (16.0 for English/Arabic and 20.0 for English/Korean). 

\textbf{Last subword is usually worse than first subword pooling:}
From Table \ref{tab:main}, we see that \textit{Last subword} pooling is clearly worse than \textit{First subword}, and is usually the lowest scoring strategy.
We hypothesize that this is because for most languages in our test set, the stem appears at the beginning of the word, meaning that the last subword will have less information content.

Arabic, however, is an exception with \textit{First subword} providing the lowest scores for all three test sets (Better Abstract, Better Phase-1 and ACE).
This is due to the fact that in Arabic, the beginning of a word commonly contains non-information bearing structures such as conjunctions, prepositions, and the definite marker \textit{Al} and the future tense prefix.
Consistent with this, we find in our morphological analysis of Arabic (detailed in the next section) that the stem of an anchor word overlaps more frequently with the last subword (69\%) than with the first subword (60\%).

\textbf{Morphology-informed pooling:}
We now explicitly show that it is important for a pooling strategy to include representations for subwords which have high semantic content.
To do this, we identify which subwords best represent a word's morphological stem---the part of a word responsible for its lexical meaning. 

It is out of scope to develop tools to automatically perform such complex morphological analysis for hundreds of languages of possible interest. 
However, there has been significant prior research in computational methods for such work in Arabic, which we leverage here as an illustrative example.
Specifically, we use \textsc{Madamira} \cite{Pasha2014MADAMIRAAF} to identify affixes or clitics at the beginning or end of an Arabic word and to delimit the central portion of the word that constitutes the stem. 
We then define a new pooling strategy (\textit{First stem} pooling) in which a word's pooled representation is derived by taking the transformer's hidden representation corresponding to the first subword which overlaps with the stem.

We preform our analysis on the Arabic test datasets.
The \textit{First stem} pooling is applied only during extraction; we chose to use the models trained on English with \textit{First subword} pooling because for English, the stem is usually at the beginning.
In Table \ref{tab:first_stem}, we see that \textit{First stem} pooling is close to the optimal strategy, highlighting that it is important for a pooling strategy to consider linguistic structures which contain high information content. 

\section{Conclusion}
This work examines the impact of subword pooling strategy on zero-shot cross-lingual information extraction, specifically focusing on the representative task of event detection.
We find that pooling can dramatically influence target language performance.
The sensitivity to pooling is shown to be determined by linguistic differences between source and the target language and how often a target-language word splits.
Because the source versus target language variations do not arise in the monolingual scenario, we show that pooling strategies do not impact monolingual cases nearly as much as they impact zero-shot cross-lingual transfer.
Of all the pooling strategies we examined, attention pooling is robust to language variations and thus adapts best to cross-lingual differences.

\section{Acknowledgements}
This research is based upon work supported in part by the Office of the Director of National Intelligence (ODNI), Intelligence Advanced Research Projects Activity (IARPA), via Contract No. 2019-19051600007. 
The views and conclusions contained herein are those of the authors and should not be interpreted as necessarily representing the official policies, either expressed or implied, of ODNI, IARPA, or the U.S. Government. 
The U.S. Government is authorized to reproduce and distribute reprints for governmental purposes notwithstanding any copyright annotation therein.

\bibliography{aaai22}

\begin{thebibliography}{20}
\providecommand{\natexlab}[1]{#1}

\bibitem[{{\'A}cs, K{\'a}d{\'a}r, and Kornai(2021)}]{acs-etal-2021-subword}
{\'A}cs, J.; K{\'a}d{\'a}r, {\'A}.; and Kornai, A. 2021.
\newblock Subword Pooling Makes a Difference.
\newblock In \emph{Proceedings of the 16th Conference of the European Chapter
  of the Association for Computational Linguistics: Main Volume}, 2284--2295.
  Online: Association for Computational Linguistics.

\bibitem[{Amrhein and Sennrich(2021)}]{amrhein-sennrich-2021-suitable-subword}
Amrhein, C.; and Sennrich, R. 2021.
\newblock How Suitable Are Subword Segmentation Strategies for Translating
  Non-Concatenative Morphology?
\newblock In \emph{Findings of the Association for Computational Linguistics:
  EMNLP 2021}, 689--705. Punta Cana, Dominican Republic: Association for
  Computational Linguistics.

\bibitem[{Basile et~al.(2014)Basile, Caputo, Semeraro, and
  Siciliani}]{Basile2014ExtendingAI}
Basile, P.; Caputo, A.; Semeraro, G.; and Siciliani, L. 2014.
\newblock Extending an Information Retrieval System through Time Event
  Extraction.
\newblock In \emph{DART@AI*IA}.

\bibitem[{Cheng and Erk(2018)}]{cheng-erk-2018-implicit}
Cheng, P.; and Erk, K. 2018.
\newblock Implicit Argument Prediction with Event Knowledge.
\newblock In \emph{Proceedings of the 2018 Conference of the North {A}merican
  Chapter of the Association for Computational Linguistics: Human Language
  Technologies, Volume 1 (Long Papers)}, 831--840. New Orleans, Louisiana:
  Association for Computational Linguistics.

\bibitem[{Conneau et~al.(2020)Conneau, Khandelwal, Goyal, Chaudhary, Wenzek,
  Guzm{\'a}n, Grave, Ott, Zettlemoyer, and
  Stoyanov}]{conneau-etal-2020-unsupervised}
Conneau, A.; Khandelwal, K.; Goyal, N.; Chaudhary, V.; Wenzek, G.; Guzm{\'a}n,
  F.; Grave, E.; Ott, M.; Zettlemoyer, L.; and Stoyanov, V. 2020.
\newblock Unsupervised Cross-lingual Representation Learning at Scale.
\newblock In \emph{Proceedings of the 58th Annual Meeting of the Association
  for Computational Linguistics}, 8440--8451. Online: Association for
  Computational Linguistics.

\bibitem[{Devlin et~al.(2019)Devlin, Chang, Lee, and
  Toutanova}]{devlin-etal-2019-bert}
Devlin, J.; Chang, M.-W.; Lee, K.; and Toutanova, K. 2019.
\newblock {BERT}: Pre-training of Deep Bidirectional Transformers for Language
  Understanding.
\newblock In \emph{Proceedings of the 2019 Conference of the North {A}merican
  Chapter of the Association for Computational Linguistics: Human Language
  Technologies, Volume 1 (Long and Short Papers)}, 4171--4186. Minneapolis,
  Minnesota: Association for Computational Linguistics.

\bibitem[{Doddington et~al.(2004)Doddington, Mitchell, Przybocki, Ramshaw,
  Strassel, and Weischedel}]{doddington-etal-2004-automatic}
Doddington, G.; Mitchell, A.; Przybocki, M.; Ramshaw, L.; Strassel, S.; and
  Weischedel, R. 2004.
\newblock The Automatic Content Extraction ({ACE}) Program {--} Tasks, Data,
  and Evaluation.
\newblock In \emph{Proceedings of the Fourth International Conference on
  Language Resources and Evaluation ({LREC}{'}04)}. Lisbon, Portugal: European
  Language Resources Association (ELRA).

\bibitem[{Fincke et~al.(2022)Fincke, Agarwal, Miller, and Boschee}]{fincke2022}
Fincke, S.; Agarwal, S.; Miller, S.; and Boschee, E. 2022.
\newblock Language Model Priming for Cross-Lingual Event Extraction.
\newblock In \emph{Proceedings of the AAAI Conference on Artificial
  Intelligence}.

\bibitem[{Huang et~al.(2022)Huang, Hsu, Natarajan, Chang, and
  Peng}]{huang-etal-2022-multilingual-generative}
Huang, K.-H.; Hsu, I.-H.; Natarajan, P.; Chang, K.-W.; and Peng, N. 2022.
\newblock Multilingual Generative Language Models for Zero-Shot Cross-Lingual
  Event Argument Extraction.
\newblock In \emph{Proceedings of the 60th Annual Meeting of the Association
  for Computational Linguistics (Volume 1: Long Papers)}, 4633--4646. Dublin,
  Ireland: Association for Computational Linguistics.

\bibitem[{Keren et~al.(2022)Keren, Avinari, Tsarfaty, and
  Levy}]{https://doi.org/10.48550/arxiv.2204.04748}
Keren, O.; Avinari, T.; Tsarfaty, R.; and Levy, O. 2022.
\newblock Breaking Character: Are Subwords Good Enough for MRLs After All?

\bibitem[{Kudo and Richardson(2018)}]{kudo-richardson-2018-sentencepiece}
Kudo, T.; and Richardson, J. 2018.
\newblock {S}entence{P}iece: A simple and language independent subword
  tokenizer and detokenizer for Neural Text Processing.
\newblock In \emph{Proceedings of the 2018 Conference on Empirical Methods in
  Natural Language Processing: System Demonstrations}, 66--71. Brussels,
  Belgium: Association for Computational Linguistics.

\bibitem[{Lan et~al.(2020)Lan, Chen, Xu, and Ritter}]{Lan2020GigaBERTZT}
Lan, W.; Chen, Y.; Xu, W.; and Ritter, A. 2020.
\newblock GigaBERT: Zero-shot Transfer Learning from English to Arabic.
\newblock \emph{arXiv: Computation and Language}.

\bibitem[{Loshchilov and Hutter(2019)}]{loshchilov2019decoupled}
Loshchilov, I.; and Hutter, F. 2019.
\newblock Decoupled Weight Decay Regularization.
\newblock \emph{arXiv:1711.05101}.

\bibitem[{Pasha et~al.(2014)Pasha, Al-Badrashiny, Diab, Kholy, Eskander,
  Habash, Pooleery, Rambow, and Roth}]{Pasha2014MADAMIRAAF}
Pasha, A.; Al-Badrashiny, M.; Diab, M.~T.; Kholy, A.~E.; Eskander, R.; Habash,
  N.; Pooleery, M.; Rambow, O.; and Roth, R. 2014.
\newblock MADAMIRA: A Fast, Comprehensive Tool for Morphological Analysis and
  Disambiguation of Arabic.
\newblock In \emph{LREC}.

\bibitem[{Pouran Ben~Veyseh et~al.(2022)Pouran Ben~Veyseh, Nguyen, Dernoncourt,
  and Nguyen}]{pouran-ben-veyseh-etal-2022-minion}
Pouran Ben~Veyseh, A.; Nguyen, M.~V.; Dernoncourt, F.; and Nguyen, T. 2022.
\newblock {MINION}: a Large-Scale and Diverse Dataset for Multilingual Event
  Detection.
\newblock In \emph{Proceedings of the 2022 Conference of the North American
  Chapter of the Association for Computational Linguistics: Human Language
  Technologies}, 2286--2299. Seattle, United States: Association for
  Computational Linguistics.

\bibitem[{Sennrich, Haddow, and Birch(2016)}]{sennrich-etal-2016-neural}
Sennrich, R.; Haddow, B.; and Birch, A. 2016.
\newblock Neural Machine Translation of Rare Words with Subword Units.
\newblock In \emph{Proceedings of the 54th Annual Meeting of the Association
  for Computational Linguistics (Volume 1: Long Papers)}, 1715--1725. Berlin,
  Germany: Association for Computational Linguistics.

\bibitem[{Toshniwal et~al.(2020)Toshniwal, Shi, Shi, Gao, Livescu, and
  Gimpel}]{toshniwal-etal-2020-cross}
Toshniwal, S.; Shi, H.; Shi, B.; Gao, L.; Livescu, K.; and Gimpel, K. 2020.
\newblock A Cross-Task Analysis of Text Span Representations.
\newblock In \emph{Proceedings of the 5th Workshop on Representation Learning
  for NLP}, 166--176. Online: Association for Computational Linguistics.

\bibitem[{Wu et~al.(2016)Wu, Schuster, Chen, Le, Norouzi, Macherey, Krikun,
  Cao, Gao, Macherey, Klingner, Shah, Johnson, Liu, Kaiser, Gouws, Kato, Kudo,
  Kazawa, Stevens, Kurian, Patil, Wang, Young, Smith, Riesa, Rudnick, Vinyals,
  Corrado, Hughes, and Dean}]{DBLP:journals/corr/WuSCLNMKCGMKSJL16}
Wu, Y.; Schuster, M.; Chen, Z.; Le, Q.~V.; Norouzi, M.; Macherey, W.; Krikun,
  M.; Cao, Y.; Gao, Q.; Macherey, K.; Klingner, J.; Shah, A.; Johnson, M.; Liu,
  X.; Kaiser, L.; Gouws, S.; Kato, Y.; Kudo, T.; Kazawa, H.; Stevens, K.;
  Kurian, G.; Patil, N.; Wang, W.; Young, C.; Smith, J.; Riesa, J.; Rudnick,
  A.; Vinyals, O.; Corrado, G.; Hughes, M.; and Dean, J. 2016.
\newblock Google's Neural Machine Translation System: Bridging the Gap between
  Human and Machine Translation.
\newblock \emph{CoRR}, abs/1609.08144.

\bibitem[{Xu et~al.(2021)Xu, Ebner, Yarmohammadi, White, Van~Durme, and
  Murray}]{haoran-etal-2022-gradual-finetuning}
Xu, H.; Ebner, S.; Yarmohammadi, M.; White, A.~S.; Van~Durme, B.; and Murray,
  K. 2021.
\newblock Gradual Fine-Tuning for Low-Resource Domain Adaptation.

\bibitem[{Yang et~al.(2003)Yang, Chua, Wang, and Koh}]{10.1145/860435.860444}
Yang, H.; Chua, T.-S.; Wang, S.; and Koh, C.-K. 2003.
\newblock Structured Use of External Knowledge for Event-Based Open Domain
  Question Answering.
\newblock In \emph{Proceedings of the 26th Annual International ACM SIGIR
  Conference on Research and Development in Informaion Retrieval}, SIGIR '03,
  33–40. New York, NY, USA: Association for Computing Machinery.
\newblock ISBN 1581136463.

\end{thebibliography}

\appendix

\section{Reproducibility Information}

\textbf{Machine configuration and software libraries:} 
All experiments were run on a single GPU, Quadro RTX 8000 (48GB). 
Software library versions are provided in Table \ref{tab:trigger_library_versions}. 

\begin{table}[h]
\centering
\begin{tabular}{@{}lllll@{}}
\toprule
Module             & Version    \\ \midrule
cudatoolkit & 10.2.89 \\
nltk & 3.4.5 \\
pip & 20.1.1 \\
python & 3.6.9 \\
pytorch & 1.6.0 \\
pytorch-crf & 0.7.2 \\
seqeval & 1.2.2 \\
spacy & 2.3.2 \\
tensorboard & 2.3.0 \\
torchvision & 0.7.0 \\
transformers & 3.5.1 \\ \bottomrule
\end{tabular}
\caption{Software library versions used for all experiments.}
\label{tab:trigger_library_versions}
\end{table}

\textbf{Hyperparameters:}
We did not perform exhaustive sweeps over hyperparameters, and the values used in our experiments and generally defaulted to values motivated by earlier experiments on similar tasks \cite{fincke2022}.

All experiments were run with the following common hyperparameters:
\begin{itemize}
    \item{optimizer: Optimization was carried out using AdamW \cite{loshchilov2019decoupled}.}
    \item{warmup proportion: Warm-up was set to 0.0.}
    \item{weight decay: Weight decay was set to 0.0.}
    \item{gradient accumulation steps: Set to 1.}
    \item{validation frequency: We perform validation on the validation set after every training epoch. The model with the best validation score (micro-f1) is used to report results on the test set.}
    \item{seeds: All scores are reported as a mean over runs initialed by 4 different seeds. We used 42, 1234 1729, 7777 as the set of seeds for all our experiments.}
    \item{training batch size: Set to 16.}
\end{itemize}

The hyperparameters which we did vary across tasks are shown below:

\begin{table}[hbt!]
\centering
\begin{tabular}{l c c c}
\toprule
Task              &  lr     & epochs & seq. length     \\ 
\midrule
  BETTER Abstract & 5e-5    & 5            & 128            \\
  BETTER Phase-1  & 5e-5    & 50           & 128            \\
  BETTER Phase-2  & 5e-5    & 50           & 128            \\
  ACE             & 5e-6    & 20           & 128            \\
  MINION          & 5e-6    & 20           & 256            \\
\bottomrule

\end{tabular}
\caption{Learning rate (lr), number of training epochs (train epochs) and maximum sequence length (seq. length) used depending on the task.}
\label{tab:hyperparameters}
\end{table}

\end{document}